\documentclass[letterpaper, 10 pt, conference]{ieeeconf}  

\IEEEoverridecommandlockouts                              

\usepackage{graphics}
\usepackage{amsmath}
\usepackage{graphicx}
\usepackage{booktabs}
\usepackage{multirow}
\usepackage{hyperref}
\usepackage{tcolorbox}
\usepackage{amssymb}
\usepackage{ragged2e}
\usepackage{bbm}
\usepackage{url}
\usepackage{marvosym}

\title{\LARGE \bf
COHERENT: \underline{Co}llaboration of \underline{He}te\underline{r}og\underline{en}eous Multi-Robo\underline{t} System with Large Language Models
}

\author{Kehui Liu$^{1,2,*,\dagger}$, Zixin Tang$^{2,3,*,\dagger}$, Dong Wang$^{2,\textsuperscript{\Letter}}$, Zhigang Wang$^{2}$, Xuelong Li$^{2,4}$, Bin Zhao$^{1,2,\textsuperscript{\Letter}}$
\thanks{$^{1}$Northwestern Polytechnical University. $^{2}$Shanghai Artificial Intelligence Laboratory. $^{3}$The Chinese University of Hong Kong. $^{4}$Institute of Artificial Intelligence, China Telecom Corp Ltd.}%
\thanks{$*$Equal contribution. $^{\dagger}$Work is done during internship at Shanghai Artificial Intelligence Laboratory. \textsuperscript{\Letter}Corresponding author.}%
}
\begin{document}
\maketitle
\thispagestyle{empty}
\pagestyle{empty}

\begin{abstract}

Leveraging the powerful reasoning capabilities of large language models (LLMs), recent LLM-based robot task planning methods yield promising results. However, they mainly focus on single or multiple homogeneous robots on simple tasks. Practically, complex long-horizon tasks always require collaboration among multiple heterogeneous robots especially with more complex action spaces, which makes these tasks more challenging.
To this end, we propose COHERENT, a novel LLM-based task planning framework for \underline{co}llaboration of \underline{he}te\underline{r}og\underline{en}eous multi-robo\underline{t} systems including quadrotors, robotic dogs, and robotic arms. Specifically, a Proposal-Execution-Feedback-Adjustment (PEFA) mechanism is designed to decompose and assign actions for individual robots, where a centralized task assigner makes a task planning proposal to decompose the complex task into subtasks, and then assigns subtasks to robot executors. Each robot executor selects a feasible action to implement the assigned subtask and reports self-reflection feedback to the task assigner for plan adjustment. The PEFA loops until the task is completed. Moreover, we create a challenging heterogeneous multi-robot task planning benchmark encompassing 100 complex long-horizon tasks. The experimental results show that our work surpasses the previous methods by a large margin in terms of success rate and execution efficiency. The experimental videos, code, and benchmark are released at \url{https://github.com/MrKeee/COHERENT}.

\end{abstract}


\section{INTRODUCTION}
\label{Sec:Intro}
\begin{figure}
\centering
\includegraphics[width=0.4\textwidth]{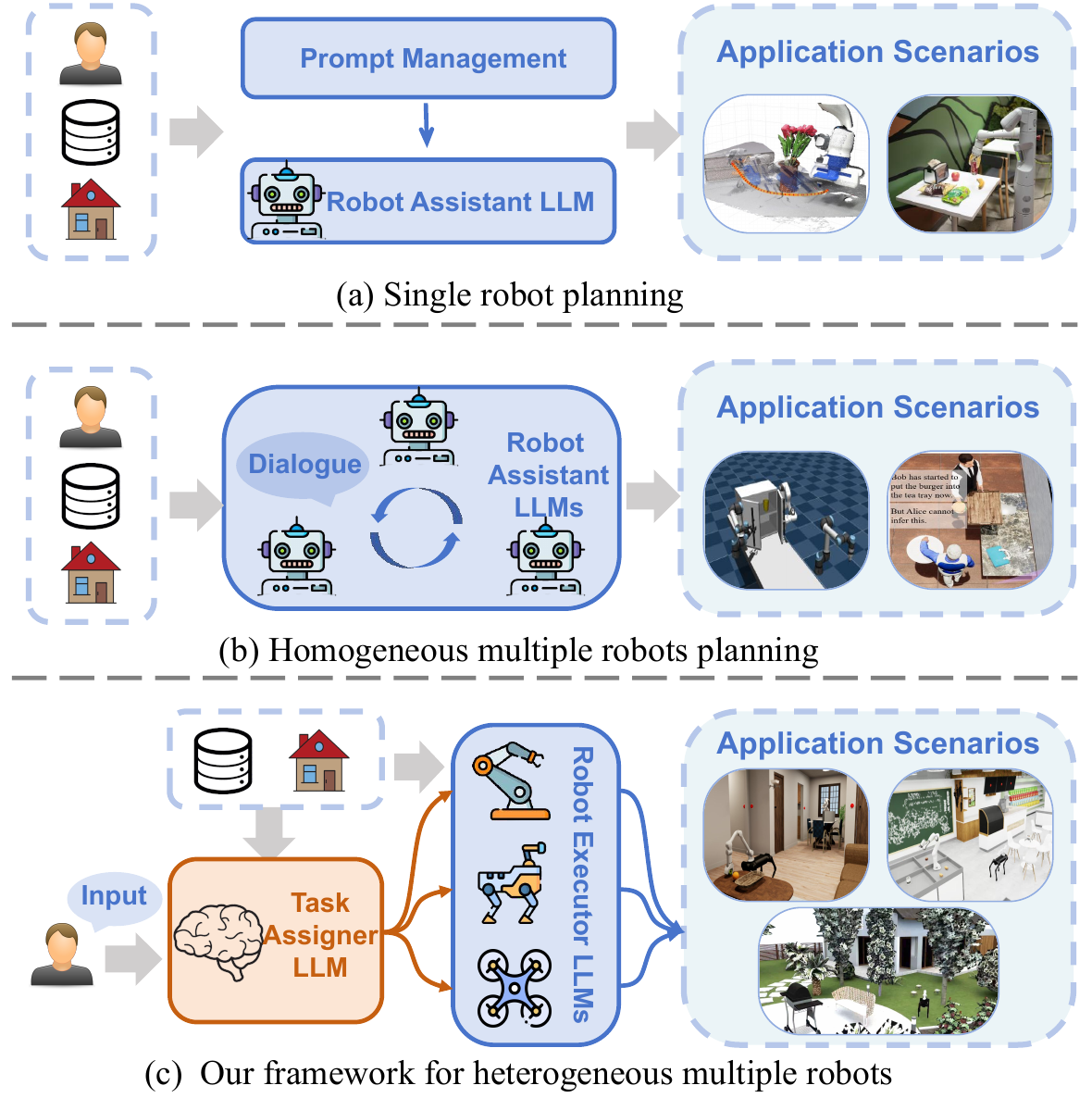}

\vskip -0.1in  
\caption{ Comparisons of LLM-based planning among (a) single-robot, (b) homogeneous multi-robot, and (c) heterogeneous multi-robot. In contrast to previous methods, \textbf{COHERENT} focuses on task allocation and collaboration among different types of robots, aiming to collaboratively accomplish complex and long-horizon tasks.}
\label{fig:intro}
\vskip -0.3in    
\end{figure}

With the rapid development of deep learning technologies in robotics, learning-based methods with extensive data produce impressive results both in single-robot \cite{shi2023robust, gu2023maniskill2} and multi-robot systems \cite{gong2023lemma}. However, even with the continuous expansion of datasets \cite{padalkar2023open, walke2023bridgedata}, the performances of these methods still struggle with generalization. The reason is that these methods lack comprehensive prior knowledge in dealing with unseen tasks, scenes and new robot types.

Encouraged by the rich world knowledge of pre-trained large language models (LLMs) and their excellent capability in complex reasoning, many recent works~\cite{kojima2022large, huang2022language, brown2020language} have utilized LLMs to analyze and decompose complex task instructions, achieving promising results in robot task planning. As shown in Figure \ref{fig:intro}(a), many single robot task planning methods leverage few-shot prompting techniques to generate multi-stage task and motion plans \cite{ding2023task}, and even robot control code \cite{huang2023voxposer, liang2023code}. Compared to single-robot systems, multi-robot systems have attracted more attention since they are able to handle more complex tasks \cite{benavidez2015design, ju2022review, queralta2020collaborative} in recent years. For homogeneous multi-robot systems, some works \cite{du2023improving, wang2023unleashing, long2023discuss, hunt2024conversational} have found that the adoption of dialogue and discussion between multiple LLM-enabled robots can release cognitive synergy ability, improve reasoning, and reduce hallucinations in complex tasks, as depicted in Figure \ref{fig:intro}(b). In \cite{mandi2024roco} and \cite{zhang2023building}, they try to generate detailed task planning via multi-turn dialogue and discussion among individual robots. However, there are many practical scenes that require collaboration between multiple heterogeneous robots, where each is characterized by different action skills. To generate correct task planning for heterogeneous robot systems, the challenge lies in accurately understanding each robot's action capabilities under changing environments, and conducting accurate task decomposition, allocation, and collaboration among different robots, which are not fully explored yet. 

To this end, we present \textbf{COHERENT}, a novel LLM-based centralized hierarchical framework for heterogeneous multi-robot task planning as shown in Figure \ref{fig:intro}(c). Our framework conducts multiple Proposal-Execution-Feedback-Adjustment (PEFA) cycles between a centralized task assigner and individual robot executors to finish a complex long-horizon task. The centralized task assigner is instantiated by prompting an LLM to decompose high-level task instructions into subtasks and assign each subtask to different robots. Each robot executor is equipped with an independent LLM and tries to accomplish the assigned subtask. Specifically, given a complex long-horizon task that requires heterogeneous robot collaboration, the task assigner first gathers all the observations and makes an initial task planning and decomposition proposal. Then, each robot tries to complete the assigned subtask by executing an action from the feasible action list of the robot, and a summarized feedback is reported to the centralized task assigner via conducting a self-reflection on whether the subtask is completed. Finally, the centralized task assigner makes an adjusted new task planning proposal based on execution feedback and historical plans to complete the overall task goal. This PEFA mechanism planning process will run in a loop until the task goal is achieved. 

To demonstrate the validity of our framework, we construct a more challenging benchmark for evaluating task planning of heterogeneous multi-robot systems than previous state-of-the-art methods. This benchmark consists of five different large-scale, realistic simulated scenes and 100 tasks with varying levels of complexity. It comprises three kinds of robots with different functions, \emph{i.e.}, quadrotors, robotic dogs, and robotic arms. Experimental results on this benchmark have confirmed the superiority of COHERENT that achieves a higher task success rate and execution efficiency. We summarize the contributions of this work as follows:

\textbf{(1)} We propose a centralized hierarchical \textbf{framework} which is more beneficial for heterogeneous multi-robot systems and enables efficient and precise task collaboration by fully leveraging robots with different action spaces. 

\textbf{(2)} We present an LLM-based proposal-execution-feedback-adjustment cyclic \textbf{mechanism} that efficiently leverages interactive feedback and adjustment to schedule different types of robots in heterogeneous multi-robot task planning.

\textbf{(3)} We construct a more challenging heterogeneous multi-robot \textbf{benchmark} containing 5 large scenes and 100 long-horizon tasks. The experimental results demonstrate the superiority of our work.

\begin{figure*}[t]
    \centering
    \includegraphics[width=0.90\textwidth]{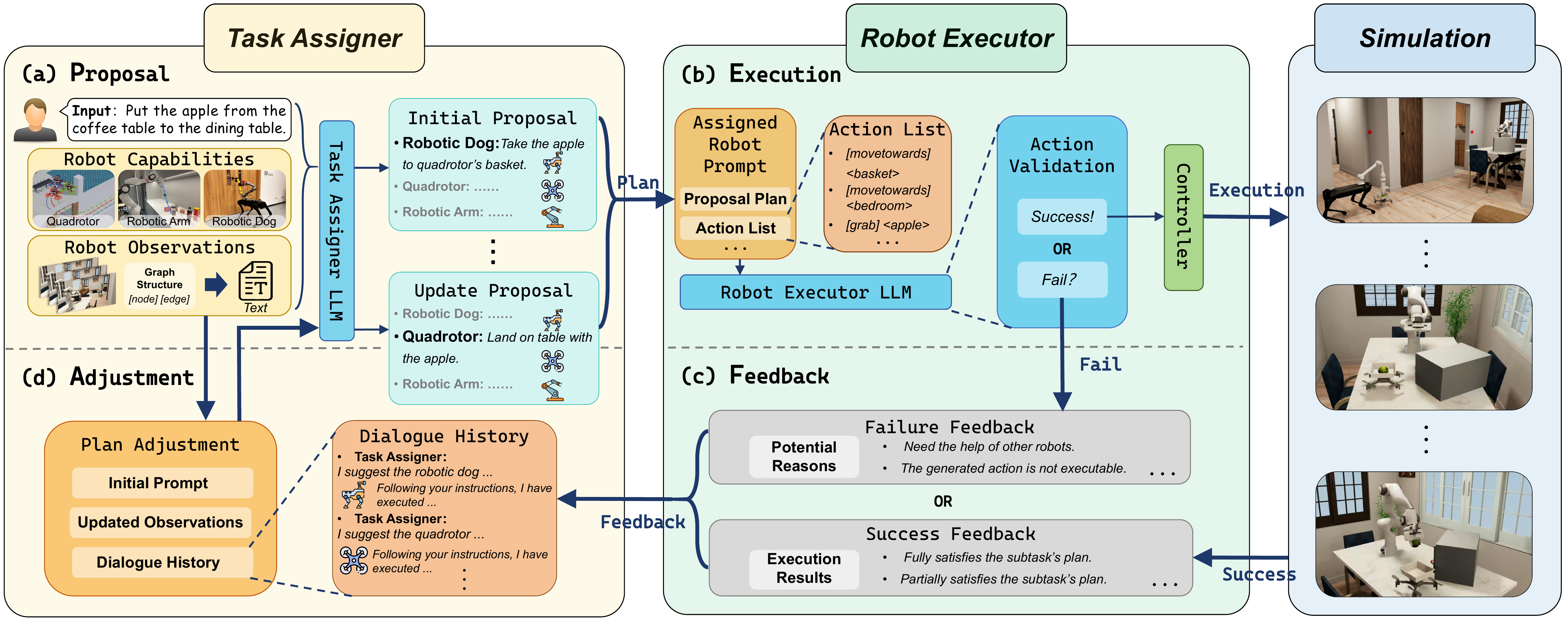}
    \vskip -0.1in  
    \caption{PEFA mechanism in \textbf{COHERENT} for heterogeneous multi-robot task planning. It consists of two stages, \emph{i.e.}, task assigner and robot executor. The proposal module decomposes the human instruction into subgoals that can be assigned to a specific robot. The execution module further maps the subgoal to an executable action. Additionally, the Feedback and Adjustment modules are designed for amendment or advancement of subtasks to complete the instruction.}
    \label{fig:pipeline}
    \vskip -0.2in  
\end{figure*}

\section{Related Works}

\subsection{Task Planning in Robotics}
Prior works can be roughly classified into three categories: classic logic-based, learning-based, and LLM-based. Classic logic-based methods \cite{aeronautiques1998pddl, hoffmann2001ff, helmert2006fast, garrett2020pddlstream}, primarily utilizing Planning Domain Description Language (PDDL) \cite{aeronautiques1998pddl}, leverage predicate logic solvers for task planning. However, these methods are unsuitable for large-scale open world and unknown environment exploration. Learning-based methods \cite{pateria2021hierarchical, nachum2018data} are represented by hierarchical reinforcement learning (HRL), which demonstrates more adaptability and flexibility in handling dynamic environments. Despite the progress made in HRL \cite{barto2003recent}, these methods continue to grapple with the well-recognized challenge of learning inefficiency, which poses substantial hurdles in their practical applicability within real-world environments. Recently, LLMs with powerful commonsense knowledge and reasoning ability present a promising avenue for addressing these limitations \cite{zhao2023large}. Several studies have been proposed to tackle various complex scenarios, such as household manipulation \cite{brohan2023can, driess2023palm, lin2023text2motion, liang2023code} and navigation \cite{long2023discuss, yu2023co}. These LLM-based methods with zero-shot \cite{kojima2022large, huang2022language} or few-shot \cite{brown2020language} prompts demonstrate robust generalization toward novel instructions and unseen environments.

\subsection{Multi-robot Planners based on LLMs}
According to the decision-making scheme, multi-robot planners based on LLMs can be categorized into two frameworks: decentralized and centralized. In dialogue-style methods \cite{mandi2024roco, zhang2023building, liu2024leveraging}, each robot delegated to an LLM agent exchanges its abilities and observations through communications to determine the next action. For example, RoCo \cite{mandi2024roco} proposes a multi-round dialogue framework with feedback to solve table-top manipulation tasks. Zhang \emph{et al}. \cite{zhang2023building} propose a modular dialogue-style framework for embodied agents to plan, communicate, and cooperate in VirtualHome-Social \cite{puig2018virtualhome, puig2020watch}. Liu \emph{et al}.~\cite{liu2024leveraging} utilizes the communication module to make ad hoc agent join the original team. However, the prompt length escalates considerably along with the number of robots involved and the iteration of dialogue \cite{chen2024scalable, mandi2024roco}, resulting in unstable performance due to LLM's inferiority in long-context reasoning. Conversely, centralized methods \cite{kannan2023smart, chen2024scalable} utilize only one central LLM to decompose and allocate the work for all robots at each planning iteration. Chen \emph{et al}. \cite{chen2024scalable} compare both dialogue-style and centralized paradigms on four multi-agent 2D task scenarios, and demonstrate that centralized communication shows a better task success rate and token efficiency than the other. Unlike most studies merely involving homogeneous settings, in this paper, we focus on designing a heterogeneous multi-robot system in large-scale realistic environments.
 
\section{Method}
\label{Sec:Method}

As shown in Figure~\ref{fig:pipeline}, \textbf{COHERENT} conducts the task assignment and robot execution in multiple loops of Proposal-Execution-Feedback-Adjustment (PEFA). We consider a living room scenario with a low coffee table and a high dining table. Three types of robots are tasked with the task instruction $\mathcal{I}$:~\textit{Put the apple from the coffee table to the dining table}. This typical heterogeneous robot collaboration task requires a quadrotor to serve as a transport bridge between the robotic dog and the robotic arm. In this section, we detail the key components of our PEFA mechanism through this example.

\subsection{Task Planning Proposal}

As shown in Figure~\ref{fig:pipeline}(a), a centralized Task Assigner LLM takes the instruction $\mathcal{I}$ as input. Meanwhile, it considers the task background, the capabilities and partial observations of all robots, historical dialogue records, and important notes within the prompt. Specifically, the task background describes our heterogeneous multi-robot settings, and each robot's capabilities are represented to distinguish different action spaces. The observations of each robot are converted into text from a graph structure, which is expressed by the \textit{relation(node1, node2)} format,~\emph{e.g.,}~\texttt{ON(apple, coffee table)}, and just include visible objects in their respective rooms. Note that each robot is unaware of environments in other rooms and objects within closed containers. Dialogue history stores memories of historical task planning proposals and execution feedback as shown in Figure~\ref{fig:pipeline}(d), and important notes are defined by some critical tips for special situations, as the fixed prompts starting with keyword \textit{Note}. The information is fused to form a long text prompt for Task Assigner LLM. The output is a task planning proposal, composed of one assigned robot and one subtask to be finished in the format~\texttt{<assigned robot>: subtask}. The complete and detailed input/output formats are shown in the supplemental videos.

In the given example, using the prompts designed above as input, Task Assigner LLM assigns specific subtasks to different robots as the initial proposal in the format: 

\begin{enumerate}
    \item \texttt{<robotic dog>: pick up the <apple> and give it to <quadrotor>}.
    \item \texttt{<quadrotor>: transport the <apple> to the <dining table>}.
    \item \texttt{<robotic arm>: put the <apple> on the <dining table>}.
\end{enumerate}
Based on the sequential order, the Task Assigner LLM outputs the first subtask to the Robot Executor LLM of the robotic dog in the next module. Each reasoning process and initial proposal will be stored in the dialogue history to update the next proposal with a different robot or plan.

\begin{figure*}[t]
    \centering
    \includegraphics[width=0.88\textwidth]{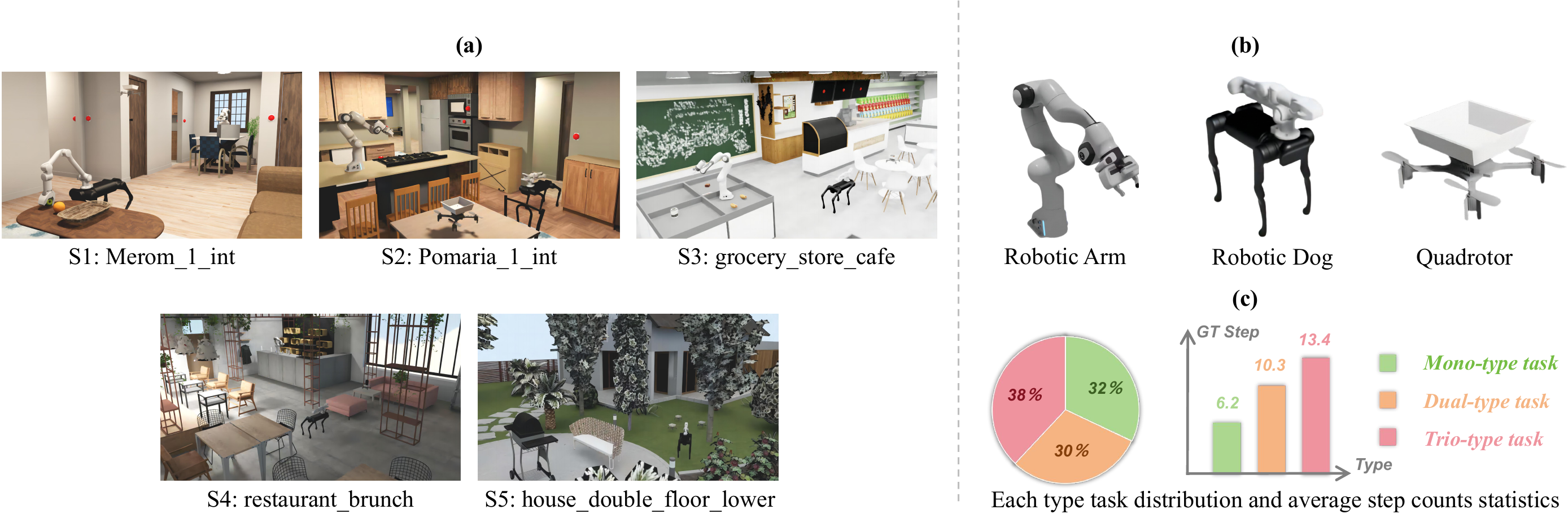}
    \vskip -0.1in
    \caption{The illustration of five scenes and three heterogeneous robots in our benchmark. In subfigure (c), the distribution of various types of tasks within the benchmark is displayed in the pie chart, along with the histogram showing the average ground truth (GT) steps for each type of task.}
    \label{fig:example}
    \vskip -0.25in
\end{figure*}

\subsection{Assigned Subtasks Execution}

\begin{table}[t]
\caption{The action list for different types of robots.}
\vskip -0.1in
\label{tab:robot_actions}
\centering
\begin{tabular}{cl}
\toprule
\textbf{Robot} & \multicolumn{1}{c}{\textbf{Action List}} \\
\midrule
Quadrotor & \begin{tabular}[c]{@{}l@{}}
{[}land\_on{]} \textless{}surface\textgreater\\ 
{[}movetowards{]} \textless{}surface\textgreater{}/\textless{}next room\textgreater\\ 
{[}takeoff\_from{]} \textless{}surface\textgreater

\end{tabular} \\
\midrule
\addlinespace 
Robotic Dog & \begin{tabular}[c]{@{}l@{}}
{[}open{]} \textless{}container\textgreater{}/\textless{}door\textgreater\\ 
{[}close{]} \textless{}container\textgreater{}/\textless{}door\textgreater\\ 
{[}grab{]} \textless{}object\textgreater\\ 
{[}putinto{]} \textless{}object\textgreater~into \textless{}container\textgreater\\ 
{[}puton{]} \textless{}object\textgreater~on \textless{}surface\textgreater\\ 
{[}movetowards{]} \textless{}object\textgreater

\end{tabular} \\
\midrule
\addlinespace
Robotic Arm & \begin{tabular}[c]{@{}l@{}}
{[}open{]} \textless{}container\textgreater\\ 
{[}close{]} \textless{}container\textgreater\\ 
{[}grab{]} \textless{}object\textgreater\\ 
{[}putinto{]} \textless{}object\textgreater~into \textless{}container\textgreater\\ 
{[}puton{]} \textless{}object\textgreater~on \textless{}surface\textgreater
\end{tabular} \\
\bottomrule
\end{tabular}
\vskip -0.2in
\end{table}

As shown in Figure~\ref{fig:pipeline}(b), the \textit{Execution} Module takes the proposal plan for the assigned robot as input and checks if the assigned subtask is executable by the assigned robot. Specifically, each assigned robot is equipped with one specific Robot Execution LLM, whose prompt is construed by the proposal plan, capabilities, partial observations and an executable action list of the assigned robot. The action list is generated based on the current state and observations using predefined rules as shown in Table~\ref{tab:robot_actions}. Note that the prompts of the different robots from the same robot type have the same capabilities but different observations and action list at each timestep. Considering the partial observation and limitations in the capability of the assigned robot, the action validation process will further determine whether it is executable in the physical world before outputting the action assigned to the robot. We assume that there is a robotic dog. For example, the action \texttt{[puton] <apple> on <dining table>} is not executable due to the height limitation. In such cases, the next \textit{Feedback} Module analyzes the failure in detail and sends the feedback to the Task Assigner. If the action is executable, it returns the task's progress to the \textit{Feedback} Module and then proceeds to the next subtask.

For the example considered, the Robot Executor LLM acts as the action selection policy to select the most appropriate action, \emph{e.g.,} \texttt{[grab] <apple>}. However, when hallucinations occur, the proposal might instruct the robotic dog to pick up the apple before it has approached it. This is an incorrect action because the apple is not within the robotic dog's manipulation range. In such cases, the robotic dog will remain still and wait for the correct instruction.

\subsection{Robot Execution Feedback}

Using the results of the action validation process as input, the self-reflection provides low-level execution feedback for high-level subtask decomposition in the next iteration, enabling bottom-up task correction. As shown in Figure~\ref{fig:pipeline}(c), this feedback includes corrections after action failures and progress updates after successful executions.

Specifically, failure feedback demonstrates the mistake in the subtask proposal or action execution. For the failure feedback, there are three typical situations. First, the proposal contains a fully wrong step to complete $\mathcal{I}$, often caused by LLM hallucinations. Second, the subtask proposals are correctly generated but assigned to the wrong robot, leading to execution issues. Third, even with the correct robot assigned, the action cannot be performed due to execution limitations. For the success feedback, the assigned robot will analyze whether the executed action fully satisfies the subtask. If the subtask requires multiple steps and the current action only partially completes it, the task’s execution progress will be updated and fed back to the Task Assigner, indicating that the subtask is not fully completed and further actions are required until the assigned robot finishes its subtask.

For example, if the robotic dog grabs the apple, it will be instructed to move toward the quadrotor as the next step. However, if the robotic dog is not near the coffee table yet, the failure feedback will indicate that it is too far from the apple to grab. In the case of an incorrect task assignment to another robot, the feedback will clarify that the robot lacks the capability to complete the task, requiring the assistance of a different type of robot.

\subsection{Subtask Proposal Adjustment}
As shown in Figure~\ref{fig:pipeline}(d), the \textit{Adjustment} Module gathers all task-level information and appends the execution feedback to the dialogue history from the most recent five iterations to improve token efficiency, which sets a solid foundation for a new PEFA cycle. This module supports the correction to the next subtask, establishing a dynamic and adaptive assigning-execution loop for the long-horizon task $\mathcal{I}$.

\section{Experiments} 
\label{sec:experiments}

\begin{table*}[t]
	\centering
	\caption{Quantitative comparison of different methods on different subsets in our benchmark.}
	\label{table:1}
    \vspace{-4mm}
	\begin{center}
	\begin{tabular}{ccccccccc}
		\toprule
		\multirow{2}{*}{\textbf{Method}}  & \multicolumn{2}{c}{\textbf{Mono-type Task}} & \multicolumn{2}{c}{\textbf{Dual-type Task}} & \multicolumn{2}{c}{\textbf{Trio-type Task}} & \multicolumn{2}{c}{\textbf{Average}}\\ 
		\cmidrule(lr){2-3}\cmidrule(lr){4-5}\cmidrule(lr){6-7}\cmidrule(lr){8-9}
		& \textbf{SR} & \textbf{AS} & \textbf{SR} & \textbf{AS} & \textbf{SR} & \textbf{AS} & \textbf{SR} & \textbf{AS} \\\hline
		
		DMRS-1D~\cite{zhang2023building} & 0.700 & 10.6 & 0.467 & 18.0 & 0.667 & 20.7 & 0.600 & 17.2     \\
		DMRS-2D~\cite{zhang2023building} & 0.500 & 11.5 & 0.267 & 19.9 & 0.400 & 24.5 & 0.375 & 19.6     \\
		CMRS~\cite{huang2022language} & \textbf{0.900} & 7.9 & 0.533 & 16.4 & 0.533 & 22.2 & 0.625 & 16.5  \\
            
        Primitive MCTS~\cite{browne2012survey} & 0.000 & 14.0 & 0.000 & 21.5 & 0.000 & 26.9 & 0.000 & 21.7  \\
        LLM-MCTS~\cite{zhao2023large} & 0.700 & 10.2 & 0.067 & 20.9 & 0.000 & 26.9 & 0.200 & 20.5 \\
        COHERENT w/o history & \textbf{0.900} & 9.0 & 0.933 & 13.9 & 0.467 & 23.9 & 0.750 & 16.5  \\

		\hline
	    COHERENT (Ours)  & \textbf{0.900} & \textbf{7.4} & \textbf{1.000} & \textbf{11.9} & \textbf{1.000} & \textbf{16.1} & \textbf{0.975} & \textbf{12.4} \\
        Ground Truth (GT) &  - & 6.5 & -  & 10.3 &  - & 12.9 &  - & 10.3  \\
		\hline
	\end{tabular}
	\end{center}
	\vspace{-5mm}
\end{table*}
\begin{table*}[t]
	\centering
	\caption{Quantitative comparison of different methods on different scenes in our benchmark.}
	\label{table:2}
    \vspace{-4mm}
	\begin{center}
	\begin{tabular}{ccccccccccccc}
		\toprule
		\multirow{2}{*}{\textbf{Method}}  & \multicolumn{2}{c}{\textbf{S1}} & \multicolumn{2}{c}{\textbf{S2}} & \multicolumn{2}{c}{\textbf{S3}} & \multicolumn{2}{c}{\textbf{S4}} & \multicolumn{2}{c}{\textbf{S5}} & \multicolumn{2}{c}{\textbf{Average}}\\ 
		\cmidrule(lr){2-3}\cmidrule(lr){4-5}\cmidrule(lr){6-7}\cmidrule(lr){8-9}\cmidrule(lr){10-11}\cmidrule(lr){12-13}
		& \textbf{SR} & \textbf{AS} & \textbf{SR} & \textbf{AS} & \textbf{SR} & \textbf{AS} & \textbf{SR} & \textbf{AS} & \textbf{SR} & \textbf{AS}  & \textbf{SR} & \textbf{AS}\\\hline
		
		 DMRS-1D~\cite{zhang2023building} & 0.500 & 17.4 & 0.625 & 15.8 & 0.625 & 18.3 & 0.750 & 15.1 & 0.500 & 19.3 &  0.600 &    17.2   \\
		DMRS-2D~\cite{zhang2023building} & 0.500 & 18.9 & 0.500 & 18.3 & 0.375 & 20.6 & 0.250 & 18.9 & 0.250 & 21.1 & 0.375  &  19.6     \\
    CMRS~\cite{huang2022language} & 0.875 & \textbf{13.1} & 0.625 & 16.6 & 0.625 & 18.5 & 0.375 & 18.1 & 0.625 & 15.9 &  0.625 &  16.5  \\
    
        Primitive MCTS~\cite{browne2012survey} & 0.000 & 21.5 & 0.000 & 21.8 & 0.000 & 22.5 & 0.000 & 20.5  & 0.000 & 22.0  &  0.000 &   21.7 \\
        LLM-MCTS~\cite{zhao2023large} & 0.250 & 20.0 & 0.250 & 20.4 & 0.250 & 21.3 & 0.125 & 19.9  & 0.125 & 20.9  &  0.200 &   20.5 \\
        COHERENT w/o history & 0.750 & 16.6 & 0.625 & 16.6 & 0.875 & 17.1 & 0.875 & 14.3  & 0.625 & 17.6  &  0.750 &   16.5 \\
        
		\hline
	    COHERENT (Ours)&\textbf{1.000}&\textbf{13.1}&\textbf{1.000}&\textbf{11.4}&\textbf{1.000}&\textbf{11.9}&\textbf{1.000}&\textbf{11.4}& \textbf{0.875} & \textbf{14.0}& \textbf{0.975 }& \textbf{12.4} \\
        Ground Truth (GT)  & - & 10.3 & - & 10.4 & - & 10.8 & - & 9.8 & - & 10.5& - &10.3  \\
		\hline
	\end{tabular}
	\end{center}
	\vspace{-5mm}
\end{table*}
\begin{figure*}[t]
    \centering
    \includegraphics[width=0.88\textwidth]{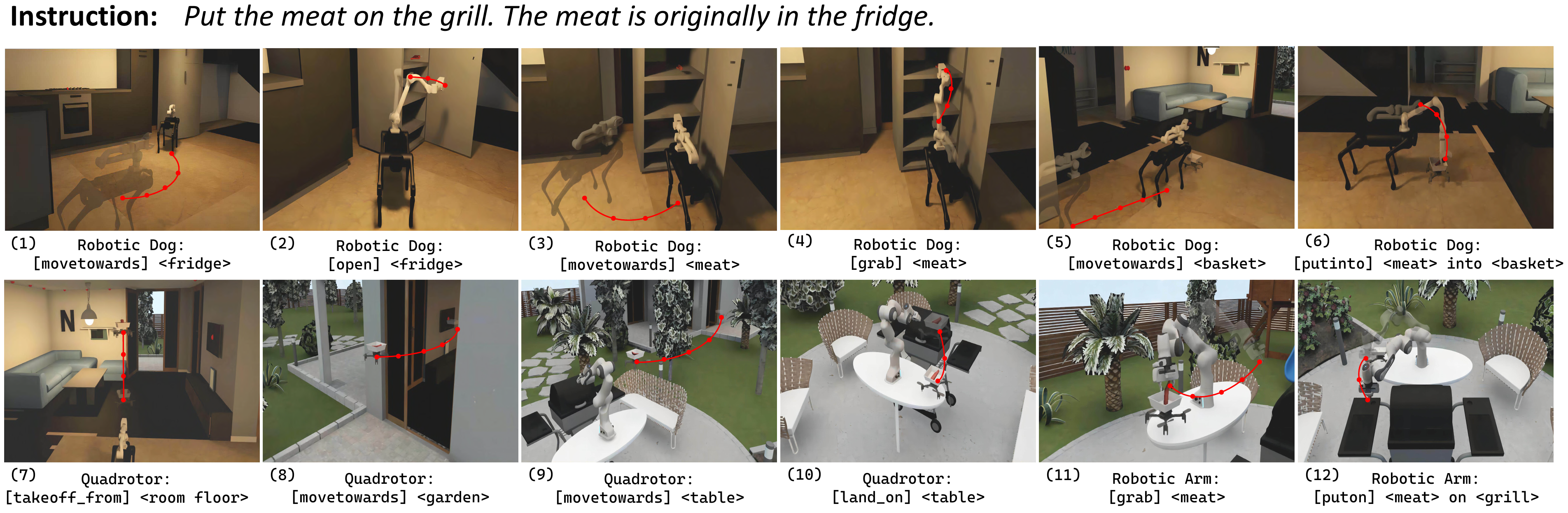}
    \vskip -0.1in
    \caption{A successful example of putting the meat on the grill in scenario S5.}
    \label{fig:example1}
    \vskip -0.25in
\end{figure*}
\subsection{Benchmark} 
As shown in Figure~\ref{fig:example}, we create a large-scale realistic embodied benchmark tailored for heterogeneous multi-robot collaboration, including quadrotors, robotic dogs, and robotic arms. Built upon the BEHAVIOR-1K platform \cite{li2023behavior}, our benchmark covers 5 typical real-world scenes: 2 apartment scenes (S1, S2), 1 apartment with garden scene (S5), 1 grocery store (S3) and 1 restaurant (S4), with a wide range of interactive objects (both rigid and articulated) and various layouts (\textit{e.g.}, multi-room and multi-floor). The ground truth (\emph{GT}) of each task represents the optimal number of steps for completion. Based on the minimum necessary number of robot types to perform each task, the benchmark is split into three categories :

\begin{itemize}
    \item \textbf{Mono-type Tasks} represent a homogeneous setting that involves only one type of robot but multiple robots of the same type are allowed. These tasks are relatively simple with \emph{GT} ranging from 4-8 steps.
    \item \textbf{Dual-type Tasks} involve the collaboration between two types of robots due to the limited capabilities of single-type robots, with a \emph{GT} step ranging from 8-12 steps.
    \item \textbf{Trio-type Tasks} show the most complex heterogeneous setting, necessitating the meticulous collaboration among all three types of robots. These tasks place high demands on the long-horizon inference capabilities of LLM-based planners, with \emph{GT} within 10-16 steps.
\end{itemize}

In total, our benchmark contains \textbf{32} mono-type tasks, \textbf{30} dual-type tasks, and \textbf{38} trio-type tasks. 

\subsection{Evaluation Metrics} 
In this paper, we adopt Success Rate (\textit{SR}) and Average Step (\textit{AS}) as the evaluation metrics. For each task, we manually design the optimal execution strategy steps as the ground truth. A task is considered successful if all objects reach their desired locations within twice the ground truth step count. For failed tasks, the execution step count is recorded as twice the ground truth plus one to facilitate uniform statistics. In summary, we aim to achieve higher success rates and fewer average steps.

\subsection{Simulation Experiments} 
\textbf{Experiment setting: }Our method is validated in the simulation environment of BEHAVIOR-1K platform \cite{li2023behavior}, employing \texttt{gpt-4-0125-preview} as LLM to conduct all experiments. Considering the cost of the OpenAI API, we select 40 tasks from all 100 tasks in the benchmark for our experiments. These tasks are sourced from all 5 distinct simulation scenes in our benchmark. 8 tasks are verified in each simulation scene, comprising 2 mono-type tasks, 3 dual-type tasks, and 3 trio-type tasks. The selected tasks allow for a comparison of our method with other baselines across various types and levels of difficulty.

\textbf{Compared methods: }Based on the different communication architectures among multiple robots, we compare our method with three other approaches. In addition to LLM-based methods, we also compare two traditional tree search algorithms in task planning to demonstrate the effectiveness of our approach.
 
\begin{itemize}
    \item \textbf{DMRS-1D: }We implement DMRS-1D baseline as a variant of CoELA~\cite{zhang2023building}, which adopts a decentralized multi-robot system framework, allowing robots to determine the next step through dialogue. The last robot summarizes all the dialogue content to reach the final conclusion.
    \item \textbf{DMRS-2D: }Considering that robots speaking earlier in a dialogue round have access to less information, we design a two-round dialogue setting to mitigate the impact of information disparity~\cite{zhang2023building}.
    \item \textbf{CMRS: }We design a primitive centralized multi-robot system~\cite{huang2022language} incorporating only a decision-making LLM. All information is stored in the prompt and then the central LLM outputs executable actions.
    \item \textbf{Primitive MCTS: }We employ the primitive Monte Carlo tree search algorithm~\cite{browne2012survey} with a random rollout strategy and the ground-truth reward function. The actions of all robots collectively form the search space.
    \item \textbf{LLM-MCTS: }We compare an algorithm that integrates LLM with the Monte Carlo method~\cite{zhao2023large} to enhance the effectiveness of primitive MCTS. LLMs are employed to serve as $\pi(a\mid h)$ in PUCT to guide action selection during the simulation process.
\end{itemize}
\vskip -0.05in
Additionally, we conduct ablation studies to evaluate the effectiveness of COHERENT without dialogue history in the prompt of Task Assigner LLM.

\textbf{Experimental Results:} We conduct experiments across various task types and different scenes in our benchmark. As shown in Table~\ref{table:1}, compared to other methods, COHERENT achieves the highest success rate and action execution efficiency. Moreover, as the difficulty of the tasks increases, the superiority of our method in terms of success rate and average step becomes more pronounced. The results of our method significantly outperform the dialogue-based decentralized method. It indicates that a global planning strategy by a central planner is crucial for task planning in heterogeneous multi-robot systems. From the results of the CMRS, we discover that dividing the process into two phases \textit{i.e.}, high-level planning and low-level action execution, allows for more precise task allocation and mapping to specific robot actions, as what our task assigner and robot executor accomplish. We also compare with the traditional tree search algorithm and its variant, and our approach still demonstrates significant advantages. In the ablation study, the absence of dialogue history in the prompt of Task Assigner LLM means that the task execution process and the feedback required to adjust the proposal plan cannot be recorded. It results in a lack of continuity for a plan that requires multiple steps to execute, thereby hindering the completion of the final goal. As illustrated in Table~\ref{table:2}, even across varied scenes, our method consistently demonstrates superior performance. This indicates its strong generalizability to different types of scenes. Figure~\ref{fig:example1} shows an example of our method successfully accomplishing the task in S5.

Furthermore, by detailing several failure reasons, we concretely analyze the problems with dialogue-based methods as a decentralized communication framework. 

\textbf{\textit{Reason 1:}} The dialogue-based method is more appropriate for robots to execute actions in parallel because each robot tends to select its own actions to achieve the objective. Lacking an understanding of other robots' capabilities and observations, it is challenging for each robot to make accurate judgments about the actions others should execute. This is the primary reason for the failure of such methods.

\textbf{\textit{Reason 2:}} Decisions made during the robot discussion process are not executed. However, subsequent robots tend to plan the next actions based on the assumption that previous actions in the dialogue have already been executed. This results in hallucinatory actions not existing in the current list of executable actions, causing failures in action parsing. An example is shown below:

\begin{tcolorbox}[left=2pt,right=2pt]
\footnotesize
\vskip -0.05in
\textbf{Subgoal:} The robotic dog is holding the apple and is close to the quadrotor. Put the apple into the quadrotor's basket to transport to the dining table (high surface).

...

\textbf{Robotic Dog:} I suggest \textless{}robotic dog\textgreater~execute {[}putinto{]} \textless{}apple\textgreater~into \textless{}basket\textgreater. (\textbf{Right})

\textbf{Quadrotor:} I suggest \textless{}quadrotor\textgreater~execute {[}takeoff\_from{]} \textless{}kitchen floor\textgreater. (\textbf{Wrong.} Because the previous action has not been executed.)

\end{tcolorbox}
\vskip -0.05in
\textbf{\textit{Reason 3:}} The actions resulting from discussions among robots are often trapped in repetitive cycles, hindering progress toward the ultimate objective. The issues stem from decision-makers lacking comprehensive information and a long-term plan. 

\begin{figure}
\centering
\includegraphics[width=0.47\textwidth]{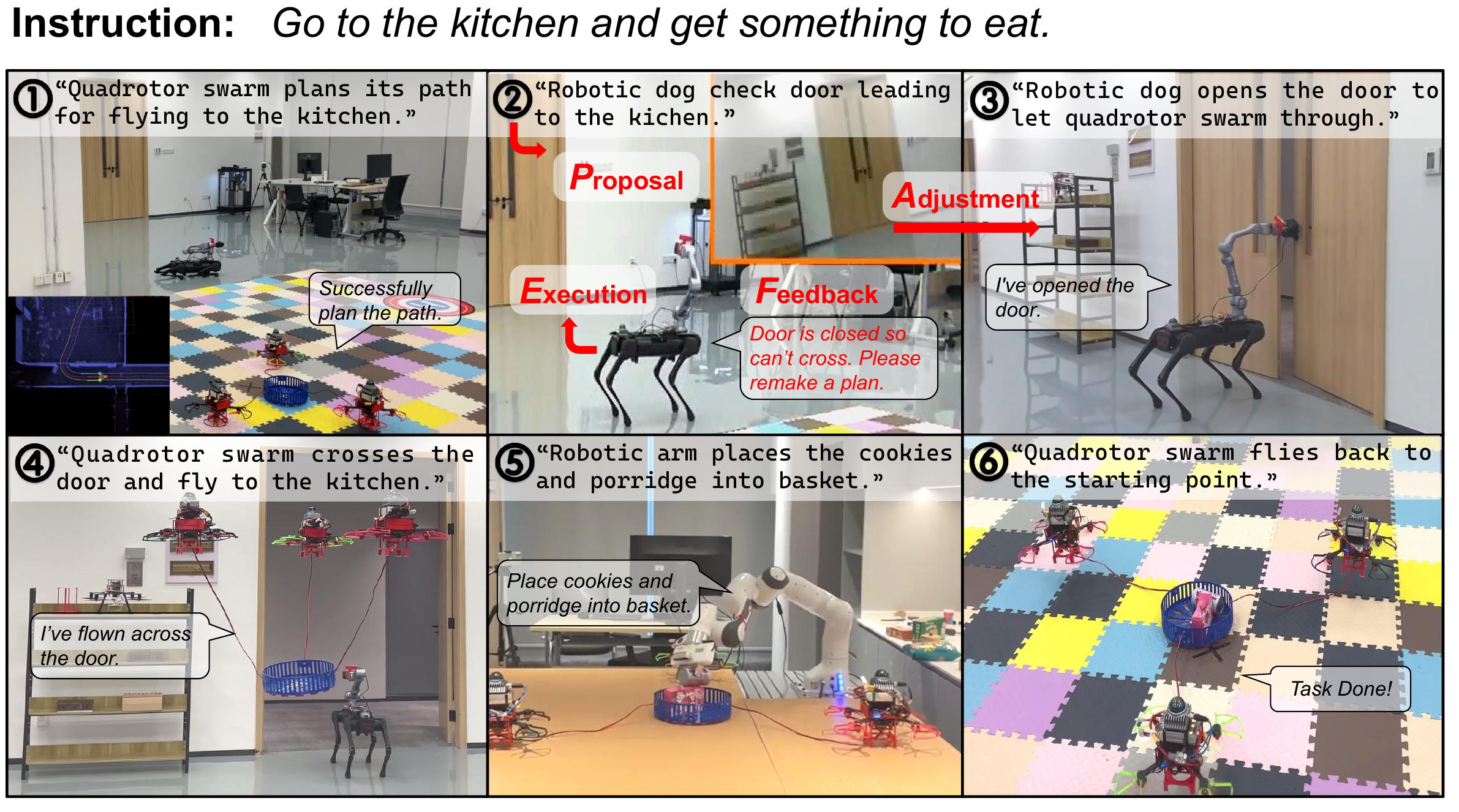}
\vskip -0.1in
\caption{\textbf{Real-world Experiment:} With the cooperation of the robotic dog and robotic arm, the quadrotor swarm fly to the kitchen across a closed door and take the cookies and porridge back which are on the kitchen table initially. This task cannot be completed without the presence of any one of the robots. }
\label{fig:exp}
\vskip -0.25in
\end{figure}

\subsection{Real-world Experiment }
To demonstrate the effectiveness of COHERENT in the real world, we conduct experiments with multiple heterogeneous robots in real-world scenes. We utilize our self-developed swarm consisting of three quadrotors, a Unitree Aliengo robotic dog equipped with a small mechanical arm on the back, and a Franka Panda robotic arm. The quadrotor swarm and robotic dog use LiDAR mapping for navigation, and the robotic arm is equipped with a wrist RGB-D camera to capture visual information. Grounding-DINO\cite{liu2023grounding} is utilized for object detection to help grasping. Besides, the communication between different robots and the host computer is based on socket connections.

As shown in Figure~\ref{fig:exp}, the user issues a voice command \texttt{Go to the kitchen and get something to eat}. The visual perception model identifies that the door to the kitchen is closed. Furthermore, the Task Assigner LLM decides to first have the robotic dog use its arm to open the door rather than let the quadrotor swarm pass the door. Then the quadrotor swarm collaborates with the robotic arm fixed in the kitchen to bring the cookies back. Our experiments prove that COHERENT can be extended to real-world environments in a zero-shot manner and exhibits generalization.
\section{CONCLUSIONS}
In this paper, we propose \textbf{COHERENT}, a novel framework designed to address task planning in heterogeneous multi-robot systems. Leveraging the reasoning capability of LLMs, we facilitate precise task allocation and timely improvements through a continuous loop of proposal-execution-feedback-adjustment between the task assigner and various robot executors. Moreover, we propose a challenging benchmark specifically tailored for heterogeneous multi-robot task planning , including 5 scenes and a total of 100 tasks across three types. Our method achieves the best results on this benchmark and has been successfully tested in the real world. 

\section*{Acknowledgments}

This work is supported by the Shanghai AI Laboratory, the National Key R\&D Program of China (2022ZD0160102), the National Natural Science Foundation of China (62376222), and the Young Elite Scientists Sponsorship Program by CAST (2023QNRC001).

\bibliographystyle{IEEEtran}
\bibliography{IEEEabrv,reference}
   
\end{document}